\documentclass[11pt]{article}

\usepackage[final]{acl}

\usepackage{times}
\usepackage{latexsym}
\usepackage[T1]{fontenc}
\usepackage[utf8]{inputenc}
\usepackage{microtype}
\usepackage{inconsolata}

\usepackage{soul}
\usepackage{url}
\usepackage{graphicx}
\usepackage{amsmath}
\usepackage{amsthm}
\usepackage{booktabs}
\usepackage{algorithm}
\usepackage{algorithmic}
\usepackage{tikz}
\usetikzlibrary{shapes.geometric, arrows, positioning}
\usepackage{tcolorbox}
\usepackage{enumitem}
\usepackage[normalem]{ulem}

\title{Separation of Skilling Context for\\Standard Operating Procedure-based Agents}

\author{Mitsuhiro Nakayama \hspace{0.5cm} Toshiaki Sota \hspace{.5cm} Takehiko Yamaguchi \hspace{.5cm} Michiaki Tatsubori\\
  IBM Japan \\
  Contact: \texttt{mich@jp.ibm.com} \\}

\begin{document}

\maketitle

\begin{abstract}
Standard Operating Procedures (SOPs) are kept concise for human readability, so LLM agents that generate executable skills from them lack the tacit knowledge and implementation details needed for automation. Human intervention fills these gaps, but must be repeated as SOPs evolve---a problem we call \textit{SOP-skill drift}. We propose managing SOPs and Contextual Knowledge (CK) separately, together with a \textit{contract-aware} evaluation that distinguishes genuine violations from CK-resolved differences. On eight Indeed-derived workflows (single LLM), it raises the functional-equivalence rate from 12\% to 58\%. Under three SOP-revision perturbations, previously extracted CK already covers 79.5\% of newly introduced automation gaps.
\end{abstract}

\section{Introduction}

Standard Operating Procedures (SOPs) are widely adopted for designing reproducible and quality-assured business workflows. SOPs document the essential steps for task execution, ensuring consistency and reliability across human operators. Recently, Large Language Model (LLM)-based agents have been increasingly used to leverage these SOPs for generating executable skills that automate complex workflows. However, a significant gap exists between high-level procedural descriptions and machine-executable logic. 

In practice, SOPs must be kept concise to ensure their readability and maintainability by human stakeholders. Consequently, they often omit granular details, creating a dependency on the ``common sense'' or tacit knowledge that human experts implicitly possess. To bridge these gaps, ad-hoc human intervention---either through direct skill editing or iterative interaction with the agent---becomes necessary to fulfill the missing requirements. This mandatory contextual knowledge is often buried within the resulting skill files, which agents use to execute the workflows.

\begin{figure}[t]
\centering
\includegraphics[width=\columnwidth]{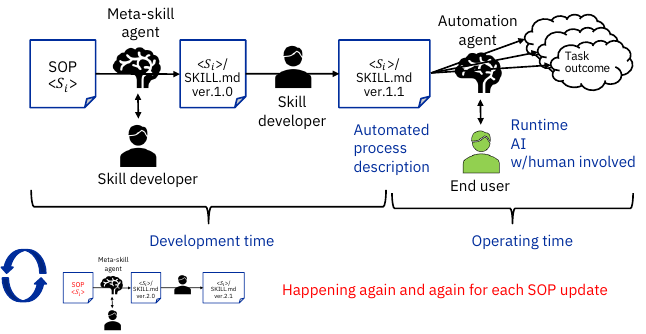}
\caption{Problem addressed in this paper: the gap between human-readable SOPs and machine-executable agent skills due to missing contextual knowledge.}
\label{fig:problem}
\end{figure}

The primary challenge arises during the evolution of business processes. As SOPs change over time, these ad-hoc human interventions must be repeated again and again to re-incorporate the necessary contextual knowledge into the updated skills. This lack of a systematic way to manage and reuse such knowledge leads to a divergence between the formal SOP and the actual skill implementation, as illustrated in Figure~\ref{fig:problem}. This drift makes subsequent maintenance increasingly difficult, as refinements are lost during skill regeneration, and the rationale behind technical adjustments remains undocumented.

To address these challenges, we propose an approach that separately manages SOPs and Contextual Knowledge (CK)---tacit knowledge and implementation details---for generating agent-executable skills. This separation keeps SOPs concise while managing environment-dependent details in a distinct layer, enabling flexible adaptation to changes. Furthermore, by using explicitly managed knowledge for skill generation, we aim to improve both reproducibility and explainability. This approach addresses our central research question: \textit{Does separately managing SOPs and implementation-level Contextual Knowledge improve the maintainability of SOP-based workflow automation systems?}

The core contribution of this work is an approach for the \textbf{explicit separation of business logic (SOPs) and implementation knowledge (Contextual Knowledge)}. We demonstrate that this decoupling improves maintainability by enabling functionally equivalent skill regeneration while significantly reducing the size of formal SOP documents. Our findings suggest that these two knowledge types can be independently managed without information loss, addressing the traditional trade-off between SOP completeness and readability. Furthermore, this separation allows changes in business logic or technical environments to be addressed in isolation, preventing the risk of SOP-skill drift.

The remainder of this paper is organized as follows: Section 2 describes the motivation and the problem of SOP evolution. Section 3 reviews related work. Section 4 presents our methodology for separately managing SOPs and contextual knowledge. Section 5 reports our experimental results. Finally, Section 6 discusses the significance of our findings and future directions.

\section{Motivation: The Challenge of SOP Evolution}

While the potential for automating Standard Operating Procedures (SOPs) with LLM agents is significant, the practical deployment of such systems faces a critical barrier: the continuous evolution of SOPs and the recurring need for human intervention. We identify two primary challenges that motivate our work.

\subsection{The ``SOP in the Wild'' Problem}
In real-world business environments, human-provided SOPs are rarely complete enough for direct automation. They often rely on the \textit{tacit knowledge} of experienced operators—contextual details about edge cases, technical environment specifics, and implicit business rules that are never formally documented. Consider a ``Standard Expense Approval'' SOP that states expenses above \$250 require manager approval. An AI agent attempting to automate this encounter a classic ambiguity: is the threshold inclusive or exclusive? Furthermore, the technical process of where to find the manager's ID or which API endpoint to call for approval is typically absent from the business-level SOP. 

\begin{figure*}[t]
\centering
\tcbset{
  skillpanel/.style={
    colframe=blue!60!black,
    colback=white,
    boxrule=0.8pt,
    arc=1pt,
    left=4pt, right=4pt, top=3pt, bottom=3pt,
    fontupper=\scriptsize,
  }
}
%
% ---- Panel 1: Source SOP ----
\begin{minipage}[t]{0.21\textwidth}%
{\small\textbf{Source SOP:}}\\[2pt]%
\begin{tcolorbox}[skillpanel]
\textbf{Obtain expense approval SOP.}\\[2pt]
The standard operating procedure for an employee to buy an item.\\[3pt]
\textbf{Purpose}\\[1pt]
The purpose of this SOP is to guarantee that all necessary approvals are obtained before purchasing an item.\\[3pt]
\textbf{Application}\\[1pt]
All employees can apply these steps to avoid any unauthorized purchase.\\[3pt]
\textbf{Procedure steps}\\[1pt]
Here are the steps of the procedure:
\begin{enumerate}[leftmargin=1.5em, topsep=2pt, itemsep=1pt, parsep=0pt]
  \item Confirm the expense is under the threshold of \$250. If it is over the threshold, the process is out of scope of this SOP.
  \item Ask the manager of the purchasing department for the approval.
\end{enumerate}
Once done, you may proceed to the Purchase Order process.
\end{tcolorbox}
\end{minipage}%
\hfill%
%
% ---- Panel 2: Generated Skill ----
\begin{minipage}[t]{0.375\textwidth}%
{\small\textbf{Generated Skill:}}\\[2pt]%
\begin{tcolorbox}[skillpanel, fontupper=\footnotesize]
\textbf{Skill: Standard Expense Approval (Under \$250)}\\[2pt]
{\color{gray}[\,\ldots\,]}\\[2pt]
\textbf{3. Operational Logic (SOP Flow)}\\[1pt]
\textbf{Step 1: Data Gathering}\\[1pt]
Before proceeding, ensure you have the following information from the user. If missing, ask for it politely.
\begin{enumerate}[leftmargin=1.5em, topsep=1pt, itemsep=0pt, parsep=0pt]
  \item \textbf{Item Name:} What do they want to buy?
  \item \textbf{Estimated Cost:} How much does it cost (in USD)?
  \item \textbf{Manager's Name:} Who is their direct approver?
\end{enumerate}
{\color{gray}[\,\ldots\,]}\\[2pt]
\textbf{4. Constraint Checklist}
\begin{enumerate}[leftmargin=1.5em, topsep=1pt, itemsep=0pt, parsep=0pt]
  \item \textbf{Never} approve an expense over \$250 using this flow.
  \item \textbf{Always} ask for the manager's name; do not assume the user acts alone.
  \item \textbf{Tone:} Professional, compliance-focused, yet helpful.
\end{enumerate}
\end{tcolorbox}
\end{minipage}%
\hfill%
%
% ---- Panel 3: Developer Edits ----
\begin{minipage}[t]{0.375\textwidth}%
{\small\textbf{Developer Edits:}}\\[2pt]%
\begin{tcolorbox}[skillpanel, fontupper=\footnotesize]
\textbf{Skill: Standard Expense Approval (Under \$250)}\\[2pt]
{\color{gray}[\,\ldots\,]}\\[2pt]
\textbf{3. Operational Logic (SOP Flow)}\\[1pt]
\textbf{Step 1: Data Gathering}\\[1pt]
\textcolor{red}{\textit{First, collect the user profile to obtain Manager's email address. Use \texttt{hr\_user\_profiles()} tool for that.}}\\[1pt]
Before proceeding, ensure you have the following information from the user. If missing, ask for it politely.
\begin{enumerate}[leftmargin=1.5em, topsep=1pt, itemsep=0pt, parsep=0pt]
  \item \textbf{Item Name:} What do they want to buy?
  \item \textbf{Estimated Cost:} How much does it cost (in USD)?
  \item \textcolor{red}{\sout{\textbf{Manager's Name:} Who is their direct approver?}}
\end{enumerate}
{\color{gray}[\,\ldots\,]}\\[2pt]
\textbf{4. Constraint Checklist}
\begin{enumerate}[leftmargin=1.5em, topsep=1pt, itemsep=0pt, parsep=0pt]
  \item \textbf{Never} approve an expense over \$250 using this flow.
  \item \textcolor{red}{\sout{\textbf{Always} ask for the manager's name; do not assume the user acts alone.}}
  \item \textbf{Tone:} Professional, compliance-focused, yet helpful.
\end{enumerate}
\end{tcolorbox}
\end{minipage}
\caption{Motivating example of human intervention: (Left) A simple SOP for expense approval; (Center) An agent-generated skill without human intervention that fails due to technical omissions and ambiguity; (Right) The developer-refined skill after human intervention through interaction with the agent or direct editing.}
\label{fig:developer_edits}
\end{figure*}

Figure~\ref{fig:developer_edits} illustrates this development gap. When an AI agent generates a skill directly from the raw SOP (center), it often misses critical implementation-level details. A developer must then manually intervene (right) to specify that the threshold is exclusive and to configure the appropriate API destinations. When an AI agent attempts to execute such an incomplete procedure without this human-in-the-loop refinement, it inevitably fails.

\subsection{The Maintenance Burden of Recurring Intervention}
A more subtle but severe problem arises as business needs change and SOPs evolve. In the expense approval example, suppose the company updates the policy to a \$300 threshold. In traditional automation workflows, the corresponding agent skill must be regenerated or modified. However, because the previously clarified details (e.g., the inclusive/exclusive rule and API configurations) were often applied directly to the implementation level without being backported to the original SOP or managed separately, these efforts are lost. This problem is exacerbated by the opaque nature of LLM-based skill generation: because the reasoning process is a black box, corrections applied at the implementation level are not automatically backported to the SOP or any structured knowledge store, causing the knowledge gap between the SOP and the skill to widen over successive iterations. 

Consequently, developers find themselves in a cycle of \textbf{recurring intervention}: they must repeatedly fix the same types of omissions and clarify the same technical ambiguities every time the SOP undergoes a minor version change. This ``SOP-skill drift'' not only increases the maintenance burden but also undermines the scalability of agentic automation. Without a mechanism to capture and refactor this contextual knowledge back into the system, the cost of maintaining an SOP-centric agent can quickly exceed the benefits of its automation.

\section{Related Work}

%\subsection{SOP-Based Agents}
Standard Operating Procedures (SOPs) have long served as the basis for business process management. Traditional approaches to automating these procedures often involve static scripting or rigid workflow engines. With the advent of LLMs, SOP-based agents have emerged, capable of interpreting natural language instructions. SOPAgent~\cite{ye2025sopagent} structures natural-language SOPs as decision graphs, explicitly managing conditional branches and long-term planning to improve task execution stability. However, these agents often rely on monolithic system prompts that suffer from context saturation and lack a systematic approach to handling procedural ambiguities. Practical frameworks like AWS Agent SOPs \cite{aws2024agentsop} provide structured workflows but still bundle business logic with execution details. More recently, \cite{giroh2025structuring} introduced a three-stage multi-agent pipeline (Clarifier, Planner, Implementor) that transforms ambiguous SOPs into executable code. The Clarifier resolves underspecification via RAG-backed Q\&A, making tacit knowledge explicit. However, this framework is designed as a one-shot batch transformation: extracted clarifications are not persisted as a reusable artifact, and SOP evolution over time is not addressed. Our approach differs in that the Contextual Knowledge (CK) layer is maintained as a persistent, versioned document, enabling accumulated knowledge to be reused across successive SOP revisions without repeated human intervention.

%\subsection*{Knowledge Management and Tacit Knowledge}
The extraction of tacit knowledge—experiential insights that are difficult to articulate—remains a bottleneck in business process automation \cite{zuin2025tacit}. Traditional BPM models often fail to capture the "how-to" expertise of human operators, leading to brittle automation \cite{nonaka1995knowledge}. Our approach addresses this by explicitly formalizing this tacit knowledge into a separate Contextual Knowledge (CK) layer, enabling more robust and maintainable automation.

%\subsection*{Architectural Decoupling and Modularity}
Recently, the paradigm of \textit{modular agent skills} has gained traction, notably within the Claude Code ecosystem \cite{Santos+:2025:ClaudeCode,anthropic2025agent_skills}. In this framework, a ``skill'' is defined as an encapsulated bundle of system instructions, executable scripts, and resources, supporting a \textit{progressive disclosure} pattern where only skill metadata is loaded at startup and the full procedural logic is injected on activation. Similarly, AWS Agent SOPs \cite{aws2024agentsop} use RFC~2119 modal verbs to specify agent behavior as structured Markdown, separating workflow logic from tool implementations. While both frameworks decouple business logic from execution infrastructure, they treat skills as static, manually authored artifacts and provide no mechanism for \textit{automatically extracting} tacit knowledge from human interventions, nor for managing the evolution of skills as the underlying SOPs change over time. Our work directly addresses this gap: rather than requiring domain experts to encode all implementation details into skill files upfront, we propose a systematic approach to capture, classify, and persist the knowledge that emerges through iterative human refinement.

Hierarchical frameworks like L2M2 \cite{geng2025l2m2} further separate high-level planning from low-level execution. MetaGPT~\cite{hong2024metagpt} realizes multi-agent collaboration for software development by leveraging SOPs, improving the quality of complex tasks through role division and structured intermediate artifacts. More broadly, meta-programming offers a pathway to automate this by treating skill logic and contextual common sense as distinct components. Our work extends this by decoupling the business logic (SOP) from implementation details (CK), directly addressing the challenge of SOP-skill drift in evolving agent systems \cite{chen2025mae}. Furthermore, SciAgents~\cite{ghafarollahi2024sciagents} demonstrates, in the context of scientific discovery automation, a design that separates foundational knowledge from context-dependent details, improving knowledge reusability and maintainability. This supports the key insight that managing ``essence'' and ``details'' at appropriate granularities contributes to the maintainability of complex workflows---directly motivating our separation of SOPs (business logic) from CK (implementation details).

%\subsection*{SOP Execution Benchmarks}
The difficulty of SOP-based automation is empirically confirmed by recent benchmarks. SOP-Bench \cite{sopbench2025} reports that state-of-the-art LLM agents achieve only 27--48\% task success on complex industrial SOPs, while SOP-Maze \cite{sopmaze2025} identifies three recurring failure modes: route blindness, conversational fragility, and calculation errors under complex contextual conditions. These results underscore that bridging the gap between human-written SOPs and reliable agent execution remains an open challenge, further motivating the need for explicit knowledge management mechanisms such as the CK layer proposed in this work.

\section{Methodology}

\begin{figure*}[t]
\centering
\includegraphics[width=\textwidth]{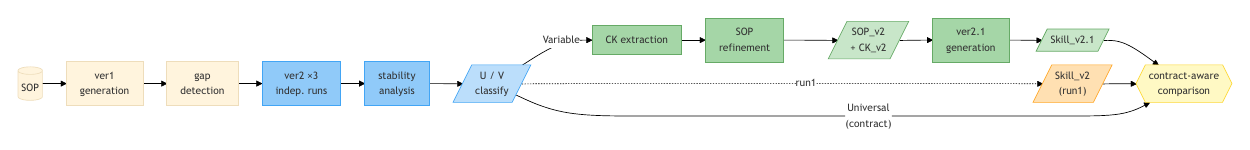}
\caption{Skill generation workflow. A developer produces Skill\_v2.md as the reference implementation via direct edits or Q\&A (bottom). The proposed approach analyzes the diff between Skill\_v1 and Skill\_v2 to extract Contextual Knowledge, refines the SOP, then regenerates Skill\_v2.1.md without further human input (top). Skill\_v2.1.md is compared against Skill\_v2.md to validate that the knowledge separation has captured all information necessary to reproduce the intended behavior.}
\label{fig:pipeline}
\end{figure*}

We propose a DevOps system that facilitates the evolution of SOPs from human-readable documents to robust, machine-executable agent skills. Our system addresses two key challenges: (1) \textbf{reproducibility}---generating consistent-quality skills from the same inputs, and (2) \textbf{explainability}---making the knowledge used for skill generation explicitly traceable. Traditional LLM-based approaches suffer from opaque reasoning processes that cannot guarantee reproducible outputs. Our core insight is that the ``missing background knowledge'' required for automation can be systematically extracted, classified, and managed as a separate layer from the SOP itself, enabling both reproducible and explainable skill generation.

\subsection{Problem Formalization}

We define three key artifacts in our framework:

\begin{itemize}
    \item \textbf{Standard Operating Procedure (SOP)}: A document $S$ describing the essential steps and business rules for a workflow. SOPs are designed for human readers and often contain implicit assumptions.
    \item \textbf{Agent Skill}: An executable implementation $K$ (including prompts, tools, and code) that enables an LLM agent to perform the workflow defined in $S$.
    \item \textbf{Contextual Knowledge (CK)}: A structured document $C$ containing implementation details and tacit knowledge (domain-specific ``common sense'') required to bridge the gap between $S$ and $K$.
    \item \textbf{Workflow Automation}: The process of making SOPs automatically executable using Agent Skills.
    \item \textbf{Human Intervention}: The process of refining Agent Skills through interaction with the agent or direct editing to fulfill knowledge gaps.
\end{itemize}

The ``SOP in the Wild'' problem can be formalized as follows: given a human-written SOP $S$, directly generating a skill $K$ often fails because $S$ lacks the granular information needed for algorithmic execution (as illustrated in Figure~\ref{fig:problem}). Our approach introduces $C$ as an intermediate layer, enabling the transformation: $S + C \rightarrow K$.

\subsection{Three-Layer Knowledge Architecture}

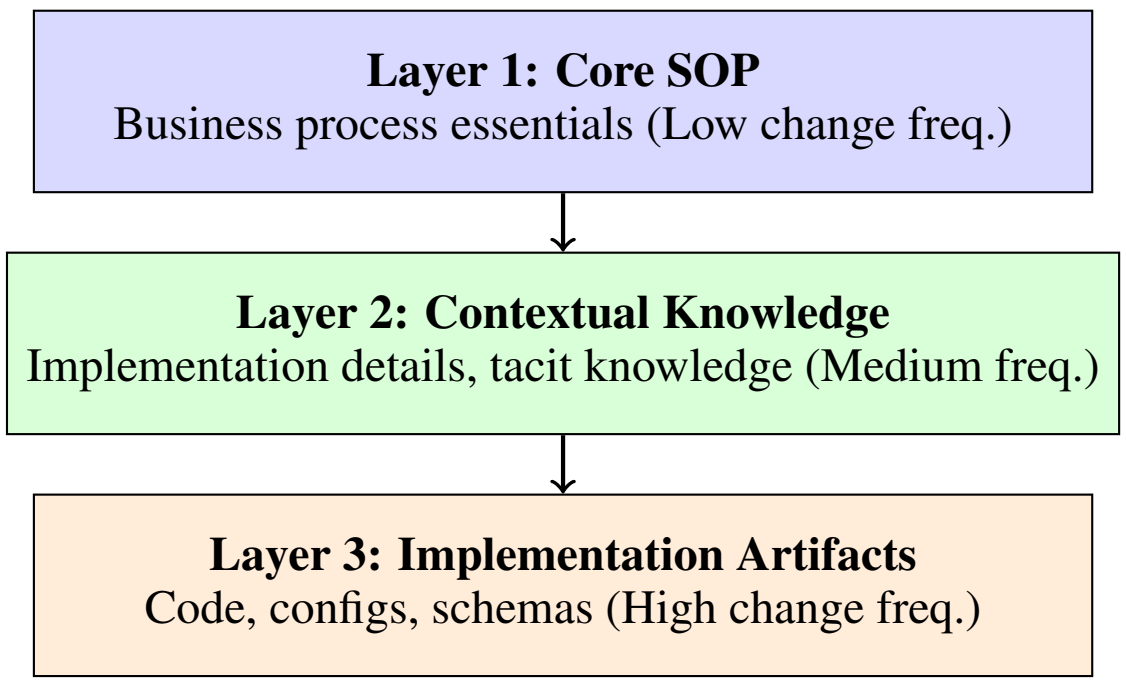
\begin{figure}[t]
\centering
\begin{tikzpicture}[
    layer/.style={draw, rectangle, minimum width=7cm, minimum height=1.2cm, align=center, font=\small},
    arrow/.style={->, thick}
]
% Layer 1: Core SOP
\node[layer, fill=blue!15] (sop) at (0, 2.4) {\textbf{Layer 1: Core SOP}\\Business process essentials (Low change freq.)};

% Layer 2: Contextual Knowledge
\node[layer, fill=green!15] (ck) at (0, 0.8) {\textbf{Layer 2: Contextual Knowledge}\\Implementation details, tacit knowledge (Medium freq.)};

% Layer 3: Implementation Artifacts
\node[layer, fill=orange!15] (impl) at (0, -0.8) {\textbf{Layer 3: Implementation Artifacts}\\Code, configs, schemas (High change freq.)};

% Arrows showing flow
\draw[arrow] (sop.south) -- (ck.north);
\draw[arrow] (ck.south) -- (impl.north);

\end{tikzpicture}
\caption{Three-layer knowledge architecture. Knowledge is separated based on change frequency and ownership to prevent SOP enlargement while maintaining consistency.}
\label{fig:architecture}
\end{figure}

Our system separates knowledge into three distinct layers based on change frequency and ownership (Figure~\ref{fig:architecture}):

\begin{enumerate}
    \item \textbf{Core SOP Layer}: Contains the essential business process logic---what to do, why, and when. This layer changes infrequently (e.g., annually) and is owned by business stakeholders.
    \item \textbf{Contextual Knowledge Layer}: Contains implementation details, technical specifications, and tacit knowledge. This layer changes at medium frequency (e.g., quarterly) and is managed by the operations team.
    \item \textbf{Implementation Artifacts Layer}: Contains executable code, configurations, and schemas. This layer changes frequently and is maintained by developers.
\end{enumerate}

This separation prevents SOP file enlargement while maintaining consistency between documentation and implementation. When a fix is applied to the skill, the system determines whether the underlying knowledge belongs to the SOP or CK layer. Crucially, by explicitly documenting tacit knowledge in the CK layer, our approach enables traceability of what knowledge was used to generate each skill, thereby improving the explainability of the generation process.

\subsection{Knowledge Separation Framework}

To classify extracted knowledge, we employ a four-category framework:

\begin{itemize}
    \item \textbf{Process Clarifications}: Implicit assumptions made explicit (e.g., ``\$250 threshold is exclusive''). These belong in the SOP.
    \item \textbf{Edge Cases}: Critical exception handling that affects workflow continuity (e.g., ``if manager is unavailable, escalate to director''). These belong in the SOP.
    \item \textbf{Domain-Specific Rules}: Business policies that users must understand (e.g., ``4-tier approval hierarchy''). These belong in the SOP.
    \item \textbf{Implementation Details}: Technical specifics such as API endpoints, library choices, and timeout configurations. These belong in the CK.
\end{itemize}

This classification ensures that the SOP remains concise and business-focused while the CK captures the technical context needed for skill generation.

\subsection{System Components}

The system architecture consists of three core components:

\textbf{Gap Detection Mechanism.} Before skill generation, an LLM-based evaluator analyzes the SOP against five criteria:
\begin{itemize}
    \item \textbf{Trigger}: Are activation conditions clearly defined?
    \item \textbf{Input}: Are all required parameters specified with validation rules?
    \item \textbf{Process}: Are execution steps unambiguous? (i.e., is the order of operations deterministic, and are all branching conditions and decision points explicitly defined?)
    \item \textbf{Output}: Is the expected result format defined?
    \item \textbf{Edge Cases}: Is error handling specified?
\end{itemize}
If any criterion is unmet, the system generates targeted questions for clarification.

\textbf{Commonsense Extractor.} This component captures human interventions through two mechanisms: (1) Q\&A interactions where users clarify ambiguities, and (2) differential analysis of user edits to generated skills. The extracted knowledge is stored in a structured \texttt{COMMONSENSE.md} artifact.

\textbf{SOP Refactoring Engine.} This module analyzes the accumulated contextual knowledge and separates it according to our four-category framework. It suggests updates to the SOP for process clarifications and edge cases, while retaining implementation details in the CK document.

\subsection{The Automation Loop}

The system operates in a continuous 4-step loop:

\begin{enumerate}
    \item \textbf{Extract}: Human fixes are captured from direct skill file edits or Q\&A interactions. The system performs gap detection and stores identified issues in \texttt{COMMONSENSE.md}.
    \item \textbf{Suggest}: Using the accumulated knowledge, the system classifies each item and proposes updates. Items classified as process clarifications, edge cases, or domain rules are suggested for SOP inclusion.
    \item \textbf{Update}: The user reviews and applies the suggested updates to the SOP (e.g., via a Pull Request in version control).
    \item \textbf{Clean}: Once updates are incorporated into the SOP, the corresponding entries are removed from \texttt{COMMONSENSE.md} to prevent redundancy.
\end{enumerate}

This loop ensures that refinement efforts are preserved and reused as SOPs evolve, differentiating our approach from traditional systems that require repeated manual intervention. Figure~\ref{fig:pipeline} illustrates the skill generation workflow, contrasting the baseline Q\&A-based approach with our proposed knowledge-driven refinement process.

Regarding the two human intervention mechanisms in the Extract step: direct skill file edits are preferred when the developer knows precisely what needs to be fixed; Q\&A is employed when the intended behavior is ambiguous and must first be elicited through iterative dialogue. In the latter case, Q\&A is more efficient than directly editing the SOP or CK, because the practitioner may not yet have a clear specification---the dialogue itself serves as a structured elicitation process that surfaces tacit knowledge before it is committed to documentation.

\subsection{Skill Creation Protocol}

For generating agent skills, we employ a 5-phase protocol:

\begin{enumerate}
    \item \textbf{Discovery}: Analyze the SOP domain and research appropriate tools and libraries.
    \item \textbf{Design}: Define use case scenarios and map human SOP steps to automated logic.
    \item \textbf{Architecture}: Plan file organization, class structure, and error handling strategy.
    \item \textbf{Detection}: Generate semantic trigger keywords and activation conditions.
    \item \textbf{Implementation}: Produce the skill artifacts including system prompts, tool definitions, and executable code.
\end{enumerate}

This structured protocol ensures consistent, high-quality skill generation that can be reproduced using the SOP and CK documents without requiring additional LLM inference to fill knowledge gaps.

\section{Experiments}

To evaluate whether separately managing SOPs and implementation-level Contextual Knowledge (CK) improves the maintainability of SOP-based workflow automation systems, we conducted experiments addressing two research questions:
\begin{itemize}
    \item \textbf{RQ1 (Comprehensiveness)}: Can the proposed system comprehensively extract implementation and tacit knowledge from human intervention? (Evaluated via Functional Equivalence and Degree of Redundancy).
    \item \textbf{RQ2 (Quality of Classification)}: Are extracted knowledge elements correctly classified as either SOP content or Contextual Knowledge? (Evaluated via Clarity of Separation and Conciseness of SOPs).
\end{itemize}

\subsection{Dataset}

We evaluated our approach on three business workflows of varying complexity. Scenario~1 was synthetically generated using an LLM to provide a controlled baseline, while Scenarios~2 and~3 were adapted from real-world SOP examples published by Indeed~\cite{indeed2024sop}, a major employment website, ensuring that our evaluation also reflects practical industry scenarios:
\begin{itemize}
    \item \textbf{Scenario 1 -- Standard Expense Approval}: An LLM-generated workflow for processing and approving expense claims with threshold-based routing (21-line SOP).
    \item \textbf{Scenario 2 -- Human Resources Hiring Flow}: A multi-step hiring procedure involving technical tests, interviews, and offer letter generation (32-line SOP).
    \item \textbf{Scenario 3 -- Copy Editing Workflow}: A complex workflow managing hand-offs between copy editors, section editors, and designers (44-line SOP).
\end{itemize}

For each workflow, we constructed a reference implementation through the following process. First, an initial Agent Skill (Skill\_v1.md) was automatically generated from the original SOP using the Skill Creation Protocol described in Section~4. Next, the Gap Detection mechanism analyzed Skill\_v1.md to identify missing information, ambiguous conditions, and unspecified edge cases. A developer then applied targeted modifications to Skill\_v1.md---informed by the gap analysis results---to produce Skill\_v2.md. This human-refined artifact serves as the ground truth for subsequent experiments, capturing the intended target behavior including implementation details, domain-specific rules, and edge-case handling that were absent from the original SOP.

\subsection{Experimental Setup}

Using the datasets constructed above, we conducted the following experimental process to address RQ1 and RQ2. Starting from Skill\_v1.md and Skill\_v2.md, the system performed four steps:

\begin{enumerate}
    \item \textbf{Contextual Knowledge Extraction}: The system analyzed the differences between Skill\_v1.md and Skill\_v2.md to extract the tacit knowledge that the developer had implicitly applied---including implementation details, domain-specific rules, and edge-case handling---and stored it as structured Contextual Knowledge (CK).
    \item \textbf{SOP Refinement}: The extracted CK was classified using the four-category knowledge separation framework. Items classified as process clarifications or edge cases were incorporated into the SOP to produce SOP\_v2.md, while implementation details were retained in the CK layer.
    \item \textbf{Skill Regeneration}: Skill\_v2.1.md was regenerated solely from SOP\_v2.md and the extracted CK---without additional human input---to verify that the separated knowledge is sufficient to reproduce the intended behavior.
    \item \textbf{Version Comparison}: Skill\_v2.1.md was compared against the ground-truth Skill\_v2.md to assess functional equivalence.
\end{enumerate}

For each of the three workflows, we conducted three independent trials of the regeneration step (Steps 3--4) to assess reproducibility. Additionally, to examine the robustness of the CK extraction process (Step 1), we performed three independent extraction runs per workflow and compared the extracted topics for consistency.

We evaluated the results along four metrics corresponding to our research questions, as detailed in the following subsection.

Full experimental artifacts including SOPs, generated skills, and source code are provided in the Supplementary Material.

\subsection{Evaluation Metrics}

We assess our system along four metrics grouped by research question.

\paragraph{RQ1 -- Comprehensiveness.}
\textit{Functional Equivalence} measures whether Skill\_v2.1.md, regenerated solely from the separated SOP and CK, reproduces the behavior of the human-refined Skill\_v2.md. This metric verifies that knowledge separation does not result in information loss: if the separated artifacts are sufficient, the regenerated skill should exhibit equivalent functionality.
\textit{Degree of Redundancy} measures whether the regenerated skill contains unnecessary additions beyond the information present in the baseline Skill\_v2.md. A high degree of redundancy would indicate that the extraction process introduces noise, undermining the maintainability benefits of separation.

\paragraph{RQ2 -- Quality of Classification.}
\textit{Clarity of Separation} assesses whether each extracted knowledge element is correctly classified as SOP content or Contextual Knowledge, evaluated through human review of the four-category classification results. Accurate classification is essential for ensuring that SOPs remain process-focused while implementation details are managed separately.
\textit{Conciseness of SOPs} quantifies whether the refined SOP\_v2.md remains compact after incorporating process clarifications, measured as the size reduction relative to the combined SOP and CK documentation. A concise SOP indicates that knowledge separation successfully prevents document bloat, keeping operational procedures accessible to non-technical stakeholders.

\subsection{Experimental Results}

We now present the results for each metric across the three scenarios.

\subsubsection{Functional Equivalence}

For each of the three workflows, we conducted three trials to generate Skill\_v2.1.md from the separated CK and SOP, then compared it against the baseline Skill\_v2.md.

\begin{table*}[h]
\small
    \centering
\begin{tabular}{lccc}
\hline
\textbf{Case} & \textbf{Trial 1} & \textbf{Trial 2} & \textbf{Trial 3} \\
\hline
Scenario 1 (Expense Approval) & FE & PE & FE \\
Scenario 2 (HR Hiring) & FE & FE & FE \\
Scenario 3 (Copy Editing) & FE & FE & FE \\
\hline
\end{tabular}
\caption{Trials of Functional Equivalence (FE: Functional Equivalence, PE: Partial Equivalence)}
\label{tab:functional_equivalence}
\end{table*}

Across all trials, the regenerated Skills were confirmed to be functionally equivalent to the original implementations in 8 out of 9 cases (89\%). The single ``Partial Equivalence'' verdict (Case 1, Trial 2) also indicated equivalent core functionality, with differences only in implementation depth---the v2.1 version included complete Python source code while v2.0 remained a specification document.

This success rate is considered practically sufficient: when a regenerated Skill does not achieve full functional equivalence, a practitioner can manually correct the output and update the Contextual Knowledge accordingly, triggering a subsequent regeneration cycle that incorporates the corrected information. This human-in-the-loop refinement loop progressively improves regeneration accuracy over iterations, making an initial success rate of 89\% an acceptable baseline for production use.

Additionally, in particular, to examine the robustness of the Commonsense Extraction process, we performed three independent extraction runs for each case and compared the extracted topics for consistency.

\begin{table*}[h]
\small
\centering
\begin{tabular}{lcccc}
\hline
\textbf{Case} & \textbf{All 3 Runs} & \textbf{2/3 Runs} & \textbf{1 Run} & \textbf{Reliability} \\
\hline
Scenario 1 & 15 topics & 2 topics & 1 topic & 93\% \\
Scenario 2 & 12 topics & 0 topics & 0 topics & 100\% \\
Scenario 3 & 9 topics & 3 topics & 1 topic & 85\% \\
\hline
\end{tabular}
\caption{Commonsense Extraction Consistency Across Independent Runs}
\label{tab:ck_consistency}
\end{table*}

\textbf{Case 1 (Expense Approval)}: All three runs independently identified 15 core topics with high consistency, including skill invocation methods (4 methods with 0.8 NLP confidence threshold), escalation timeline (Day 3/5/7/10/15), backup approver hierarchy (4 levels), request lifecycle states (7 states), and duplicate detection thresholds (80\% text similarity, 5\% amount tolerance). Minor variations existed in topic granularity (14--16 topics per run) but no contradictions were found.

\textbf{Case 2 (HR Hiring)}: The extraction process demonstrated excellent reliability with 100\% topic agreement. All 12 core topics appeared in all three runs: personnel directory configuration, system integration specifications, technical test configuration, business calendar handling, candidate withdrawal procedures, rescheduling policies, evaluation disagreement resolution, offer negotiation framework, system failure recovery, concurrent candidate management, extended status tracking, and email template standardization. Content agreement rate was approximately 95\% with zero significant contradictions.

\textbf{Case 3 (Copy Editing)}: Nine topics appeared universally across all runs with strong content agreement, covering authorization, input validation, workflow management, and safety mechanisms. Three additional topics appeared in two of three runs (issue severity classification, workflow transition rules, cross-article consistency checks), indicating these may be considered secondary knowledge.

\subsubsection{Degree of Redundancy}

Using Skill\_v2.md as the baseline containing necessary information, we measured the size increase in Skill\_v2.1.md and examined the additional content.

\begin{table*}[h]
\small
\centering
\begin{tabular}{lccc}
\hline
\textbf{Case} & \textbf{Skill\_v2} & \textbf{Skill\_v2.1 (Range)} & \textbf{Increase} \\
\hline
Scenario 1 & $\sim$1,057 lines & 2,315--2,716 lines & 119--157\% \\
Scenario 2 & $\sim$1,282 lines & 2,456--3,237 lines & 92--153\% \\
Scenario 3 & $\sim$672 lines & 858--1,495 lines & 28--122\% \\
\hline
\end{tabular}
\caption{File Size Comparison Between Skill Versions}
\label{tab:file_size}
\end{table*}

The size of Skill\_v2.1.md increased compared to Skill\_v2.md. However, detailed analysis revealed that the additional content consisted of: (1) complete Python implementation code, (2) deployment artifacts (requirements.txt, marketplace.json, README.md), (3) technology stack decisions and rationale, and (4) project scaffolding and file structure specifications.

These additions represent necessary implementation guidance rather than redundant elements. We observe that, while Skill\_v2.md excels as a business requirements document with comprehensive configuration examples and interaction scenarios, Skill\_v2.1.md excels as a technical implementation guide with working code and project scaffolding.

\subsubsection{Clarity of Separation}

Human evaluators assessed the classification results to determine whether each element was appropriately categorized as SOP content or CK.

\begin{table*}[h]
\small
\centering
\begin{tabular}{lccc}
\hline
\textbf{Classification} & \textbf{Scenario 1} & \textbf{Scenario 2} & \textbf{Scenario 3} \\
\hline
Process Clarifications & 4 & 5 & 5 \\
Domain-Specific Rules & 3 & 3 & 3 \\
Edge Cases & 4 & 4 & 4 \\
Implementation Details & 6 & 3 & 3 \\
\hline
\textbf{Total} & \textbf{15} & \textbf{12} & \textbf{13} \\
\hline
\end{tabular}
\caption{Knowledge Classification by Category}
\label{tab:classification}
\end{table*}

The evaluation revealed that boundary conditions (e.g., ``\$250 threshold is inclusive or exclusive'') tended to be appropriately classified as Process Clarifications requiring SOP updates, while technical specifications (e.g., API endpoint designs, library choices) were correctly retained as Implementation Details in the CK.

We observe the classification step is effective using our knowledge separation framework (Process Clarification, Domain-Specific Rules, Edge Cases, Implementation Details), as follows:
\begin{itemize}
    \item Can clarify what should be added to SOP\_v2: Process Clarification, Edge Cases
    \item Can be separated as implementation guide: Implementation Details
    \item Can distinguish organization-specific rules: Domain-Specific Rules
\end{itemize}

\subsubsection{Conciseness of SOPs}

We measured the file size of the separated SOP relative to the combined SOP and CK documentation.

\begin{table*}[h]
\small
\centering
\begin{tabular}{lcccc}
\hline
\textbf{Case} & \textbf{Original} & \textbf{SOP\_v2} & \textbf{Combined} & \textbf{Reduction} \\
\hline
Scenario 1 & 21 lines & 72 lines & $\sim$446 lines & 84\% \\
Scenario 2 & 32 lines & 110 lines & $\sim$498 lines & 78\% \\
Scenario 3 & 44 lines & $\sim$95 lines & $\sim$380 lines & 75\% \\
\hline
\end{tabular}
\caption{SOP Size Metrics}
\label{tab:sop_size}
\end{table*}

Across the three workflows, the separated SOPs achieved a 75--84\% reduction in file size compared to the combined SOP and CK documentation. While the enhanced SOP\_v2 files grew compared to original SOPs (to incorporate essential process clarifications and edge cases), they remained significantly more concise than if all contextual knowledge had been embedded directly.

Key additions to SOP\_v2 files included:
\begin{itemize}
    \item \textbf{Scenario 1}: Scope clarification (\$250 boundary), required information fields, timeline expectations, modification/cancellation procedures
    \item \textbf{Scenario 2}: Business day definitions, role responsibilities, disagreement resolution procedures, rescheduling policies, withdrawal handling
    \item \textbf{Scenario 3}: Stage-specific processing instructions, role authorization matrix, escalation handling procedures
\end{itemize}

\subsection{Discussion}

%\subsubsection*{Summary}

The experiments provide affirmative answers to both research questions. For RQ1 (Comprehensiveness), the system achieved 89\% functional equivalence (8/9 trials) with 100\% core business logic preservation, and the Commonsense Extraction process demonstrated 85--100\% reliability across independent runs. For RQ2 (Classification Quality), the four-category framework correctly classified knowledge elements, with process clarifications and edge cases elevated to SOP content while implementation details remained in the CK layer. The separation approach achieved 75--84\% SOP size reduction compared to combined documentation, supporting improved maintainability.

These results validate the effectiveness of our three-layer knowledge architecture: separating Core SOP, Contextual Knowledge, and Implementation Artifacts based on change frequency and ownership keeps SOPs concise while CK captures the tacit knowledge necessary for reproducible skill generation. The four-category classification framework provides clear guidance for determining whether extracted knowledge belongs in the SOP or CK layer, and Gap Detection serves as an effective quality gate by proactively identifying missing information before skill generation.

\subsubsection{Validity and Limitations}

Regarding validity, our evaluation relied on human judgment with LLM assistance for functional equivalence determination (internal validity), was limited to three workflows from a single domain source using a single LLM (external validity), used proxy metrics that may not capture practical maintainability value without user studies (construct validity), and depends on LLM capabilities that may change over time (reproducibility).

Key limitations include:
\begin{enumerate}
    \item \textbf{Scale of Evaluation}: Three workflows with three trials each provides initial evidence but is insufficient for statistical significance claims. Larger-scale evaluation across diverse industries would strengthen confidence.
    \item \textbf{Automation and Knowledge Decay}: The current implementation requires human review at several stages. We did not address mechanisms for detecting when CK becomes stale or contradicts updated SOPs.
    \item \textbf{Workflow Complexity}: The evaluated workflows are relatively linear. Workflows with extensive branching, parallel execution, or cross-departmental handoffs may present additional challenges.
\end{enumerate}

\subsubsection{Future Work}

Future research directions include:
\begin{enumerate}
    \item \textbf{Automated Classification}: Extending the knowledge separation framework with machine learning models trained on larger SOP-CK corpora, and investigating multi-agent orchestration for complex workflows.
    \item \textbf{Longitudinal Studies}: Tracking SOP-CK co-evolution over extended periods to provide empirical evidence for maintainability benefits, combined with user studies for human-in-the-loop optimization.
    \item \textbf{Generalization}: Evaluating cross-LLM generalization and domain-specific adaptations for highly regulated industries such as healthcare and finance.
\end{enumerate}

\section{Conclusion}

We proposed an approach that separately manages SOPs and Contextual Knowledge (CK) for generating agent-executable skills. Our three-layer knowledge architecture separates Core SOP, Contextual Knowledge, and Implementation Artifacts based on change frequency and ownership, keeping SOPs concise while enabling flexible adaptation to environmental changes.

Experiments on three business workflows---expense approval, hiring process, and copy editing---demonstrated that our approach can successfully extract buried contextual knowledge from human intervention without introducing significant redundancy. The resulting SOPs achieved a 75--84\% reduction in file size compared to the combined SOP and CK documentation, while preserving functional equivalence. These results provide an affirmative answer to our research question: separately managing SOPs and Contextual Knowledge improves the maintainability of SOP-based workflow automation systems by preventing SOP-skill drift.

In particular, our approach is effective in the following respects:
\begin{itemize}
    \item Tacit knowledge buried in skill files is made explicit as CK, improving the reproducibility of skill generation.
    \item SOP readability is preserved while all necessary knowledge is managed separately, avoiding SOP file bloat.
    \item The recurring problem of repeating the same corrections at each SOP update is resolved, reducing maintenance costs.
    \item The independence of business logic and implementation knowledge is ensured, improving resilience to changes in either dimension.
\end{itemize}

Future work includes extending our approach to more complex multi-step workflows, investigating automated methods for knowledge classification, and integrating with broader enterprise workflow orchestration platforms.

\section*{Limitations}
Our evaluation has several limitations. Regarding validity, functional equivalence was determined via LLM-as-Judge with contract-aware prompts (internal validity), the evaluation covers eight workflows from a single source using a single LLM (external validity), and proxy metrics may not fully capture practical maintainability value without user studies (construct validity).

Key limitations include: (1)~\textbf{Scale of Evaluation}---eight workflows with three trials each (24 total) strengthens confidence over our initial 3-workflow study but remains insufficient for statistical significance claims; larger-scale evaluation across diverse industries would further strengthen results. (2)~\textbf{Evaluation Independence}---the contract-aware and naive judgments use the same LLM (Claude Opus) with different prompts, so judge independence is not strictly guaranteed. (3)~\textbf{Classifier Calibration}---the four-category classifier (Process / Domain / Edge / Implementation; see Appendix Table~\ref{tab:classification}) reported no misclassifications across all eight workflows, but the absence of any flagged misclassification is suggestive of a \textit{ceiling effect}: the current judge prompt may not stress-test enough failure modes to produce mistakes. We therefore treat the 100\% correctly\_placed metric as a ceiling-effect outcome rather than positive evidence of calibration, and re-evaluation with adversarial prompts is warranted. (4)~\textbf{U/V Ratio Observation}---the pattern that U/V~$\geq$~1.0 coincides with regeneration failures is based on only two failure cases (n=2); it is reported as an exploratory observation rather than a primary contribution and requires additional validation on a larger corpus, particularly given that the U/V~$=$~0.94 case (Customer Service) achieved triple-FE, indicating the boundary at U/V~$=$~1.0 is fragile. (5)~\textbf{Universal vs.\ SOP requirement}---our Universal classification reflects LLM output stability across runs, not human-validated SOP requirements. The mapping between the two is assumed but not measured. (6)~\textbf{Automation and Knowledge Decay}---we did not address mechanisms for detecting when CK becomes stale or contradicts updated SOPs.

One planned validation experiment produced only partial evidence within the submission window owing to recurrent inference endpoint instability (intermittent socket disconnects and 1800--3600\,s timeouts on long-output prompts): the contract-aware version\_comparison cross-judge re-evaluation completed 0/24 trials before being deferred (each trial took $\sim$32~min and failed mid-stream); only the lighter naive-prompt cross-LLM judgment completed (Appendix~\ref{sec:appendix_xllm}). The no-CK ablation completed 19/24 trials (the five missing trials---all of Case~9, Case~10/try~3, Case~11/try~1---hit a systematic 3600\,s endpoint timeout on the no-CK ver2.1 generation step despite three retry attempts each); the 19 completed trials provide enough signal to rule out the trivial-absorption hypothesis (Appendix~\ref{sec:appendix_ablation}), but a full 24/24 sweep on a more reliable endpoint remains future work.

A no-report ablation (Appendix~\ref{sec:appendix_noreport}) empirically quantifies the information leakage concern: withholding the U/V report from the judge reduces FE from 58\% to 29\% (one-sided McNemar $p{=}0.046$). CK alone still doubles the naive rate (+17pp), but the U/V report contributes a further +29pp. The measured +46pp contract-aware improvement thus reflects both genuine CK value and an amplification effect from U/V labels---future work should explore judge designs that use CK without requiring upstream U/V classification.

Future research directions include validating the U/V ratio hypothesis on a larger corpus of SOPs (particularly cases with U/V~$\geq$~1.0), completing the contract-aware cross-LLM validation and the missing 5 no-CK ablation trials on a more reliable inference endpoint, evaluating cross-LLM generalization, and conducting longitudinal studies tracking SOP-CK co-evolution over extended periods.

\section*{Ethical Statement}

\textbf{Data and Privacy.} The SOP examples used in our experiments were adapted from publicly available resources and do not contain personal or sensitive information. No human subjects were involved in the data collection process.

\textbf{Societal Impact.} While workflow automation can improve efficiency and reduce human error, we acknowledge its potential impact on employment. Our approach maintains human oversight by requiring explicit review and approval of SOP updates, ensuring that domain experts remain integral to the automation process. We encourage organizations adopting such technologies to consider responsible deployment strategies that support workforce transition and skill development.

\textbf{Reproducibility and Transparency.} A key motivation of this work is to improve the reproducibility and explainability of LLM-based skill generation. By explicitly managing Contextual Knowledge separately from SOPs, our approach enables traceability of the knowledge used in automation, contributing to more transparent AI systems.

\textbf{Use of Generative AI.} Generative AI tools (Anthropic's Claude models, including \texttt{claude-opus-4-7} and \texttt{claude-sonnet-4-6}) were used during manuscript preparation to assist with drafting and revising prose, restructuring sections (including the long-to-short paper conversion), integrating experimental results into the manuscript, and refining, correcting, and editing text. All AI-generated content was reviewed and verified by human authors, who remain fully responsible for the content of this work, including all factual claims, experimental design and results, and methodological choices.

\bibliography{bibs/sop}

%\clearpage
%\input{sections/appendix}

\end{document}